\documentclass[11pt,letterpaper]{article}

\usepackage{graphicx}
\usepackage{amsmath}
\usepackage{amssymb}
\usepackage{epsfig}
\usepackage{graphicx}
\usepackage{color,soul}
\usepackage{amsfonts}
\usepackage{fullpage}
\usepackage{appendix}
\usepackage{epsfig}
\usepackage{authblk}

\usepackage{algorithmic}
\usepackage{algorithm}
\usepackage{color}

\title{Compositional Structure Learning for Action Understanding}

\author[1]{Ran Xu\thanks{rxu2@buffalo.edu}}
\author[1]{Gang Chen\thanks{gangchen@buffalo.edu}}
\author[2]{Caiming Xiong\thanks{caimingxiong@ucla.edu}}
\author[1]{Wei Chen\thanks{wchen23@buffalo.edu}}
\author[3]{Jason J. Corso\thanks{jjcorso@eecs.umich.edu}}
\affil[1]{Department of Computer Science and Engineering, SUNY at Buffalo}
\affil[2]{Department of Statistics, UCLA}
\affil[3]{Department of Electrical Engineering and Computer Science, University of Michigan}

\begin{document}

\maketitle

\begin{abstract}
The focus of the action understanding literature has predominately been classification, however, there are many applications demanding richer action understanding such as mobile robotics and video search, with solutions to classification, localization and detection. In this paper, we propose a compositional model that leverages a new mid-level representation called compositional trajectories and a locally articulated spatiotemporal deformable parts model (LALSDPM) for fully action understanding. Our methods is advantageous in capturing the variable structure of dynamic human activity over a long range. First, the compositional trajectories capture long-ranging, frequently co-occurring groups of trajectories in space time and represent them in discriminative hierarchies, where human motion is largely separated from camera motion; second, LASTDPM learns a structured model with multi-layer deformable parts to capture multiple levels of articulated motion. We implement our methods and demonstrate state of the art performance on all three problems: action detection, localization, and recognition.
\end{abstract}

\section{Introduction}
\label{sec:intro}

Classifying human actions in video, commonly called \textit{action recognition} 
in the literature, has received wide attention over the last decade.  Advances 
in both features 
\cite{OritWolfECCV2012,WaKlScCVPR2011,KlMaScBMVC2008,LaMaScCVPR2008} and 
representations 
\cite{SadCorsoCVPR2012,WaUlKlBMVC2009,NiChFeECCV2010,YangGregCVPR2009} coupled with more challenging datasets such as UCF50/101 
\cite{ReShMVA2012,SoZaShTR2012} and HMDB51 \cite{KuJhGaICCV2011} have led to an 
unforeseen action classification capability.  Novel and socially enriching 
applications such as video search with semantic action indexing instead of strictly 
low-level feature indexing \cite{JiHaXiACMMM2012} are around the corner. 

However, many potential applications of action recognition in video require 
more than just action classification.  For example, unconstrained human-robot 
interaction \cite{ArBiARS2010} requires localization of action; natural 
language video description requires full detection, localization and 
classification of action to generate rich text, unlike current methods that have been 
able to do with only classification \cite{DaXuDoCVPR2013,KrMaMoAAAI2013}.  

Yet, relatively few works have emphasized these important aspects of 
action understanding---solutions to action localization, detection and 
classification.  Most early works are based on rigid, manually chosen templates 
\cite{ArronJamesPAMI2001,GoBlShTPAMI2007,DeSiCaCVPR2010}, or deforming models 
\cite{NiChFeECCV2010,YaoZhuICCV2009} that miss joint space-time deformation 
(see Sec. \ref{sec:related} for a longer review).  

More recently, a space-time deformable parts model (SDPM) was proposed by Tian 
et al.  \cite{TiSuShCVPR2013} that can capture space-time articulation for full 
action understanding.  But, this model is limited: first, as a direct extension from state of the art object detection method \cite{FeGiMcPAMI2010}, the cuboid-nature of the parts and the two layer star model render them limited in modeling the rich structural, kinematic and dynamic variability of human motion \cite{JoPP1973}. Second, it depends on a weak underlying feature (HOG3D) \cite{KlMaScBMVC2008}, which is shown to be less powerful than HOG/HOF \cite{LaMaScCVPR2008} in representing the variation in human action.

A second line of promising work for action understanding is based on point 
trajectories.  Originally proposed by Messing et al. \cite{MePaKaICCV2009}, 
point trajectories capture motion articulation in space-time and when coupled 
with rich descriptors like HOG/HOF \cite{LaMaScCVPR2008} and are densely computed 
\cite{KlMaScBMVC2008}, achieves state of the art performance for action 
classification. A limitation of the dense trajectories are that they are 
short-lived and limited in modeling the full extent of an articulated; another limitation is that they may fall on moving background rather than human action.  
Furthermore, grouping trajectories seems promising in capturing relationships between various articulating action parts, Raptis et al. \cite{RaptisSoattoCVPR2012} recently made a step in this direction to overcome above limitations by clustering trajectories. But, in their model the location of the structures is fixed before learning, therefore limiting the generality of the approach.

\begin{figure}[t]
\centering
\includegraphics[width=1\linewidth]{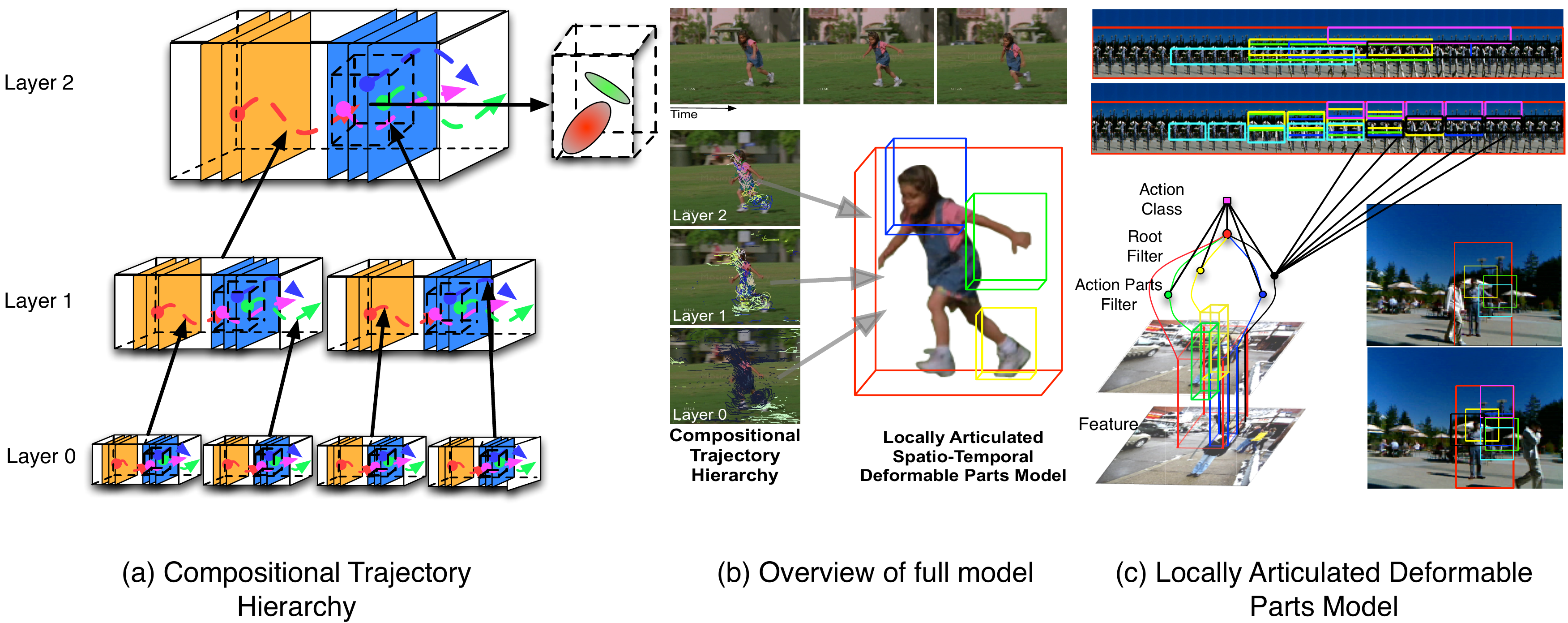}
\caption{\small{(b) illustrates our compositinoal structured model with two components. (a) illustrates the representation of compositional trajectory in three layers where elements are composed within spatio-tempooral neigborbood, as dashed cube shows. Accumulation of compositions forms local maxima with spatial-temporal distribution shown by colored ellipsoid. (c) illustrates our multi-layer parts model, the upper two images show real inference results by SDPM \cite{TiSuShCVPR2013} and LASTDPM, the parts by SDPM (upper) is of rigid shape, while the parts by LASTDPM is deformable, noticing the up and down of yellow subparts tracking motion of ``handwaving". The lower left image shows the graphical model of LASTDPM and lower right images show root and part/subpart location in certain frames.}}
\label{fig:overview}
\end{figure}

As discussed by Chen et al. \cite{WeiCorsoCVPR2014}, motion in a video can occur in various forms such as agent (human/animal) moving, camera panning or jittering, background object moving, among many others. We are particularly interested in human action understanding, where a video can be decomposed into human action and other motion, then human action can be decomposed into articulated body parts with motion and appearance, and further decomposed into articulated sub-parts and so on. We observe that actions of different classes, such as ``moving arm" in running and walking, share many common and recurring elements, and when those articulated elements merge together we further obtain highly discriminative, long-range action parts. In order to model the compositionality of human action from low-level representation to high-level semantic action parts, we propose a compositional model in two steps: (1) we learn a compositional hierarchy based on co-occurring statistics; (2) given the hierarchical representation, we learn a structured model with multiple layers of parts (see Fig.  \ref{fig:overview} (b) for overview of the two steps).

In the first step of our model, we adopt a bottom-up approach and propose a new mid-level representation called \textit{compositional trajectories}. The basic idea is that we learn a hierarchical compositional model that starts with dense trajectories as the basic elements and then recursively groups frequently co-occurring pairs of elements.  At higher levels, the composed trajectories focus on the salient action parts (filtering is a byproduct) and discriminative articulations among action parts hierarchies (see Fig.  \ref{fig:overview} for an example of three layers in the hierarchy). The new representation itself has already outperforms a complex Markov random field over trajectory grouping \cite{RaptisSoattoCVPR2012} in action localization without bounding box annotation for training (see Sec. \ref{exp:localization} for detail).  

In the second step, as Fig. \ref{fig:overview} (c) illustrates, we learn a structured model called locally articulated spatiotemporal deformable parts model or LASTDPM. Our model is based on the learned compositional hierarchies and a three-layer deformable parts hierarchy, which enables us to capture the global articulation of an action with parts 
that are more locally discriminative compared with \cite{TiSuShCVPR2013}, as demonstrated by our action recognition and detection results in Sec. \ref{exp:recognition} and Sec. \ref{exp:detection}.

\section{Related Work}
\label{sec:related}



In this section, we discuss recent advances in action recognition, localization and detection. 

\paragraph{Action Recognition}
Recently, researchers have focused on developing better video feature and representation. Representative low-level features include HoG3D \cite{KlMaScBMVC2008}, HOG/HOF \cite{KlMaScBMVC2008}, dense trajectory \cite{WaKlScCVPR2011} and its variants \cite{jain2013better,WangSchimidICCV2013} . Middle-level representations that utilize human pose \cite{AngelaLucBMVC2011,XuCorsoAMDO2012,WangYuilleCVPR2013} provide a different angle to the problem and demonstrate compensative to low-level features. High-level representations such as Action Bank \cite{SadCorsoCVPR2012} introduces action space and carry rich semantic meaning. More recently, deep learning \cite{KarpathyLiCVPR2014} is applied for large-scale action recognition. 

\paragraph{Action Localization}
Given a video with human action, localization answers the question of when and where the action happens. In \cite{RaptisSoattoCVPR2012}, salient spatiotemporal structures form clusters of dense trajectories \cite{WaKlScCVPR2011} are detected as candidates for the 
parts of an action; a graphical model captures spatiotemporal dependencies and 
is used  to infer the action localization. Note that the location of salient 
structures is fixed before learning the graphical model, unlike in our case 
which jointly learns both. Lan et al. propose a figure-centric model \cite{TianMoriICCV2011} for joint action localization and recognition, while the localization is based on bounding 
box of human detection, and implicitly enforces temporal constrains between 
neighboring frames, but they assume figure is fully visible for the entire 
duration of video. Ma et al. \cite{MaSclaroffICCV2013} propose a new representation called hierarchical space-time segments for action recognition and localization, which leverages the power of hierarchical segmentation in frame level. 



\paragraph{Action Detection}
Action detection holds no assumption of given video and answers the question of whether, when and where certain action happens. A line of works detect action by explicit template matching process. The global template can be explicitly constructed 
\cite{DeSiCaCVPR2010,ArronJamesPAMI2001,GoBlShTPAMI2007,EfrosMalikICCV2003,YuanWuPAMI2011}, 
or estimated from many exemplars \cite{RoAhShCVPR2008}.  These methods all have 
rigid templates, but recent work has emphasized non-rigid templates such as Ke 
et al. \cite{KeHebertICCV2007} which divides the global template into 
independent parts and then integrates their scores for matching---note that the 
parts in their work are supervised unlike in our method which are latent---and Yao et 
al. \cite{YaoZhuICCV2009} that capture an action as a sequence of frame 
exemplars. Another line of works explore the notion of \textit{parts}, Niebles et al. \cite{NiChFeECCV2010} extend part from spatial segment to a set of consecutive video frames, but their method can only detect action temporally; SDPM \cite{TiSuShCVPR2013} directly extend DPM to space-time domain, but the part structures from their two layer model are initialized in a data-driven manner. 


\section{Learning the Compositional Hierarchies}
\label{sec:comptraj}

We design compositional trajectories as a hierarchy of spatiotemporally flexible 
compositions that characterize both articulated motion and embedded appearance 
information. Our compositional model is inspired by the work of Fidler and 
Leonardis \cite{FiLeCVPR2007}, which learns a compositional model for 
objects based on statistical co-occurrence of oriented Gabors. 
In Sec. \ref{sec:ct:def} we define the building blocks in our model, which we 
call compositions, and then a frequency-based scheme is applied to learn the 
statistically most significant compositions in each layer of the hierarchy described in Sec. \ref{sec:dt:learn}. Given a testing 
video, Sec. \ref{sec:dt:infer} introduces an efficient way to infer compositions.

\subsection{Definition}
\label{sec:ct:def}

Human action has a high degree of articulation. To distinguish large 
intra-variance of the same action, the representation should encode enough 
flexibility spatial-temporally; to benefit from similar motion patterns of 
distinct actions, the representation should be shareable in lower layer of the 
hierarchy; and to make the composition of parts distinguishable, parts should 
bear strong motion and appearance information. Our representation satisfies all 
three of these desiderata.  

We initialize the first layer using point trajectories \cite{MePaKaICCV2009}, 
due to their spatiotemporal flexibility over rigid cuboids \cite{DoRaCoPETS2005} 
and their increased descriptiveness over sparse points \cite{LaIJCV2005}.  
Motivated by the success of dense trajectories \cite{WaKlScCVPR2011} in action 
classification, we leverage dense trajectories as the basic building blocks in 
layer 0. 

Denote $L_n$ as the $n$th layer, each element\footnote{For the compositional 
trajectories, we use the term \textit{elements} (or \textit{compositions}) instead of \textit{parts} as was used in \cite{FiLeCVPR2007} to distinguish them from the different 
\textit{parts} we define in Sec. \ref{sec:lastdpm}.} in $L_n$ is a composition of 
sub-elements (i.e. elements from previous layer). Let $P_i^n$ be $i$-th element in 
$n$-th layer. We use a simple 3D spatiotemporal spring deformation model to  
capture the spatial and temporal relation of $P_i^n$ and its sub-elements. Consider 
$P_i^n$ in the center of a cube (i.e., located at $(0,0,0)$ and encompassing a 
list ${(P_j^{n-1},(x_j,y_j,t_j),(\sigma_{1j},\sigma_{2j},\sigma_{3j}))}_j$, 
where $(x_j,y_j,t_j)$ denote the relative position of $P_j^{n-1}$ and 
$(\sigma_{1j},\sigma_{2j},\sigma_{3j})$ denote variance of its position around 
$(x_j,y_j,t_j)$. (See Fig. \ref{fig:overview} (a) for illustration.) With all above information, each element can be identified by a unique id, which we call \textit{element type}. In each layer, we define a set of upward links, denoted as $Link_n$, that maintain a list of all parts of $L_{n+1}$ that $P_i^n$ indexes to for fast inference (see Sec. \ref{sec:dt:infer}). 

To initialize $L_0$ with dense trajectory, we sampled $n$ trajectory descriptors from training videos and build a codebook with $m$ visual words. Each 
trajectory that is computed directly from the video is an element in layer 
$L_0$ and the element type is the codebook index to which it best matches. In 
contrast, \cite{FiLeCVPR2007} define layer 0 elements as one of a small number 
of oriented Gabor filters in 2D. Although suitable for object shape, our 
approach allows more flexibility to handle the variability present in articulated 
action. Though the structure of $L_0$ is fixed, we learn all of the elements 
in the rest of the hierarchy automatically.

\subsection{Learning the Compositional Trajectories}
\label{sec:dt:learn}

Learning the hierarchy of compositional trajectories aims at finding 
statistically significant combinations of trajectories, in terms of motion 
compatibility, appearance compatibility and relative spatiotemporal location. 

Consider a hierarchy learned up to layer $n$.
For each element in $L_n$, we consider each element (referred as the central 
element) in the center of a cube with size $(2*r+1,2*r+1,2*l+1)$ where $r$ is 
spatial radius and $l$ is temporal radius. Since our elements are composed 
trajectories, we regard the last point temporally in the center of this cube. 

Given the central element, we seek to discover the spatiotemporal configurations of 
other local elements in $L_n$. Assume element type size in $L_{n-1}$ is $s$, 
thus a spatial-temporal map with size of $(2*r+1)^2*(2*l+1)*s^{2}$ is 
maintained. During the learning process, for each element in each video, we 
store a 3D map that accumulates the frequency of all elements in $L_n$ that have their first sub-element located within the cube to encode spatiotemporal relation of two elements. 
For each one of $s^2$ element type combinations, we find $N$ significant 
compositions after performing 3D local maxima in each such 3D map. Then, we 
generate the spatiotemporal relation $(x_j,y_j,t_j)$ and corresponding variance 
$(\sigma_{1j},\sigma_{2j},\sigma_{3j})$, as illustrated in Fig. \ref{fig:overview} (a) by dashed cube and colored ellipsoid. We consider those significant 
compositions to be candidate elements for $L_{n+1}$ and select compositions with 
highest frequency as elements in $L_{n+1}$ after inference. 

To allow for element sharing across classes, we jointly learn the compositions 
using videos from all classes at both layers $L_0$ and $L_1$.  At higher levels 
we use class-specific videos and hence learn class-specific compositions.

\subsection{Detection of Elements in Videos}
\label{sec:dt:infer}

Given a video, we initialize elements in $L_0$ by generating dense trajectory descriptor and encode each element type using codebook from training videos. According to the $Link$ we stored in training process, we can link back to compositions, e.g. $P^n$ in higher layer from current trajectories. Then we check whether there is a spatial-temporal match between current trajectory and sub-elements of $P^n$, by checking the location deformation. 

\noindent\textbf{Classification and Localization with Compositional Trajectories}\quad
Once the compositional trajectories of a video are extracted, we can directly 
use these as mid-level features for action classification.  As a basis of 
comparison, we simply use a bag of compositional trajectories, but other more 
sophisticated methods for using the compositional trajectories are possible. For action localization, we follow the same setting as \cite{RaptisSoattoCVPR2012} and simply take the spatiotemporal region of compositional trajectories as human action regions. 

\section{Locally Articulated Spatio-Temporal DPM}
\label{sec:lastdpm}

For action detection, we need to know all the detailed information of where, when and what action happens in the video. For this purpose we propose a multi-layer deformable parts model using compositional trajectories as the mid-level part descriptor; it uses histograms over their elements and allows the domain of the histograms to locally articulate in space-time for adapting to the variation in an given action class. 

\subsection{Define Sub-parts}

Fig. \ref{fig:overview} (c) shows the grachical model of LASTDPM where an action can be detected as a bounding subvolume, shown as the red bounding box in the upper images. SDPM \cite{TiSuShCVPR2013} define parts as a cubic subvolume which captures a relatively long range of the part motion and dynamics, but its ability to locally deform to handle small articulations such as bending limbs is compromised due to its rigid shape over time. To handle dynamics like this, we introduce \textit{sub-parts} (see small blocks of the second top image in Fig. \ref{fig:overview} (c)) to incorporate locally articulation action parts. We divide each cubic subvolume into $m$ subvolumes, allowing those subvolumes to spatially deform in order to fit the local motion. The subvolumes serve as the domains for our histogram accumulator on the compositional trajectory features.  We jointly learn root filter, part filters with spatial-temporal 
deformation and subpart filters with local deformation, and obtain action parts 
with deformable shape. 

Formally, for a LASTDPM with $n$ parts and $m$ subparts per part, the model is 
defined by $(2n+2(m*n)+2)$-tuple $(F_0, \{(P_{i},\{SP_{ij}\}_m \}) \}_n)$ where $F_0$ represents root filter, each $P_i$ models the $i$-th part and $SP_{ij}$ models the $j$-th subpart of the $i$-th part. Refering to Fig. \ref{fig:overview} (c) bottom-left, we define part $P_i$ by 2-tuple $(F_i,d_i)$, where $F_i$ is the part filter for $i$-th part, noting that $v_i = (v_{iy},v_{ix},v_{it})$ is a three dimensional vector indicates the anchor position of part $i$ relative to the root position, $d_i$ is a six dimensional vector that weighs the deformation cost for each possible placement of part relative to anchor position. In the third layer, we define subpart $SP_{i,j}$ by 2-tuple $(F_{sub}(i,j),d_{sub}(i,j))$, where $F_{sub}(i,j)$ is $j$-th subpart filter for $i$-th part and $d_{sub}(i,j)$ is four dimensional local deformation weights accordingly. The part scores are derived by the 3D generalized distance transform and subparts score are derived by 2D generalized distance transform because its local deformation is only spatial. We only allow local articulation in 2D to make action parts compact and capture significant and consistent motion (and our experiment clearly demonstrate the added benefit of the spatially deforming subparts). We score an action hypothesis as follows:
\begin{align}
 &score(p_0,\{p_i,\{sp_{i,j}\}_{j=1}^{m}\}_{i=1,}^{n})
= \sum_{i = 0}^{n} F_i\cdot\phi(H,p_i) +\sum_{i=1}^{n}\sum_{j=1}^{m}F_{sub}(i,j)\cdot\phi(H,sp_{i,j})
\nonumber\\
&\quad\quad\quad\quad\quad\quad\quad\quad - \sum_{i = 1}^{n} d_i\cdot\phi_{d_{3d}}(d_{x_i},d_{y_i},d_{t_i} ) -\sum_{i = 1}^{n}\sum_{j = 
1}^{m}d_{sub}(i,j)\cdot\phi_{d_{2d}}(d_{x_{ij}},d_{y_{ij}} )+ b \enspace.
\end{align}
where deformation features of parts and subparts are $\phi_{3d}(d_{x_i},d_{y_i},d_{t_i} ) = 
(d_{x_i},d_{y_i},d_{t_i},d_{x_i}^2,d_{y_i}^2,d_{t_i}^2)$ and $\phi_{2d}(d_{x_{ij}},d_{y_{ij}} ) = 
(d_{x_{ij}},d_{y_{ij}},d_{x_{ij}}^2,d_{y_{ij}}^2)$. Note that $H$ is the feature map of our compositional trajectory hierarchies, we quantize first three layers of compositional trajectories to form the grain level and quantize layer 0 and layer 1 to form the fine level features in the map.

\subsection{Inference and Training}

We design a two stage inference method for LASTDPM: (1) Localize root and part location with 3D distance transform; (2) relocalize subpart location by applying the 2D distance transform based on part location from first stage. The advantage of this formulation is that it can track motion in a locally articulated manner.  Fig. \ref{fig:overview} (c) illustrates the difference between LASTDPM and SDPM by showing how the local articulations deform to capture the idiosyncrasies of the activity. We are aware that we could use dynamic programming to infer all layers of parts, but the computational complexity will be $O(W^2H^2T)$ and sometimes makes training in large-scale data set a burden, while our inference method keeps the complexity $O(WHT)$, same as two layer case \cite{TiSuShCVPR2013}. ($W$, $H$ and $T$ are width, height and temporal length of the feature map). 

Training the LASTDPM affords the same latent SVM framework with the inference replaced with this two stage method. Denote $w  = (F_i, F_{sub}(i,j), d_i, d_{sub}(i,j), b)$ as all the parameters in the model, we train $w$ from labeled examples $<x_i, y_i>$ where $x_i$ being the vdieo and $y_i = (y_i^l, y_i^b)$ being the annotation containing class label $y_i^l$ and bounding subvolume $y_i^b$. Each example with latent variable $z$ can be classified with a function by the form of
\begin{align}
f_w(x) = \max_{z\in Z(x)} w\cdot\Phi(x,z) \enspace
\end{align}
Then the objective function is
\begin{align}
M_{D}(w)= \frac12||w||^2 + C\sum_{i=1}^{n} \max\bigl(
0,1-y_i f_w(x_i)
\bigr)
\end{align}
where $C$ controls the regularization term. The optimization problem is solved by using stochastic gradient descent, and we relabel positive samples and mine hard negative samples during training as \cite{FeGiMcPAMI2010}.


\section{Experimental Results}
\label{sec:experiments}

Our experimental setup surveys the three action understanding problems: 
recognition, detection and localization. We use challenging 
datasets with different scenarios and compare to state of the art methods. For space, we made an attempt to choose data that can be evaluated across more than one of the 
problems where possible.

\subsection{Datasets and Experiments Setup}
\noindent\textbf{UCF Sports Dataset}\quad
UCF Sports dataset \cite{RoAhShCVPR2008} consists of 150 videos captured in 
realistic scenarios with complex and cluttered background showing a large 
intra-class variability. It includes ten actions: swinging, diving, kicking, 
weight-lifting, horse-riding, running, skateboarding, swinging, golf swinging 
and walking. And it provides the frame-level annotations, we create the bounding 
volume based on the annotations for a given action video. In our paper, we adopt 
Lan et al's \cite{TianMoriICCV2011} experimental methodology on the UCF sport 
dataset, we split the data into disjoint training (103 videos) and testing 
(47 videos) set.  UCF Sports is used in all three problems.

\noindent\textbf{HOHA Dataset}\quad
Hollywood1 Human Action (HOHA) \cite{LaMaScCVPR2008} dataset has been collected 
from Hollywood movies. It contains 430 videos with eight actions: AnswerPhone, 
GetOutCar, HandShake, HugPerson, Kiss, SitDown, SitUp, StandUp. In our 
experiment, we use the clean training set. In total, there are 219 video 
sequences for training and 211 video sequences for testing. 
HOHA is used for action localization.

\noindent\textbf{Experimental Setting}\quad
We test our methods on three tasks: action recognition, action localization and 
detection. First we extract the compositional trajectories in each video 
sequence, note that in $L_0$ we sampled 100000 trajectory descriptors from training videos and build a codebook with 100 visual words. Then we set up the following steps for the three action tasks. For action recognition, we simply adopt a bag-of-words representation of the compositional trajectories in each video and use the well-known libsvm toolbox 
\cite{ChLiTlST2011} to train classifiers. For action localization, following the evaluation method in \cite{RaptisSoattoCVPR2012}, we calculate the localization score for our compositional trajectories that is $\frac{1}{|V|\cdot T}\sum_{i=1}^{|V|}\sum_{t=1}^T[\frac{|D_{i,t}\bigcap L_t}{D_{i,t}}\geq \theta]$. $L_t$ is the set of points inside the annotated 
bounding box, $[\cdot]$ is the zero-one indicator function, $D_{i,t}$ is the 
set of points belonging to the trajectories and $\theta$ is a threshold 
defining the minimum ratio of trajectories that considers it as a part of the 
bounding box. Finally, for action detection, we train our LASTDPM model with 
our compositional trajectories as the core features. We employ the common 
``intersection-over union'' criterion and generate the ROC curve for overlap 
criterion as 0.2 and also show the ROC curve for different overlap criteria by 
the area-under-curve (AUC) measure. 

\subsection{Comparative Quantitative Results}

\begin{figure}[t]
\begin{center}
\includegraphics[width = 0.65\textheight]{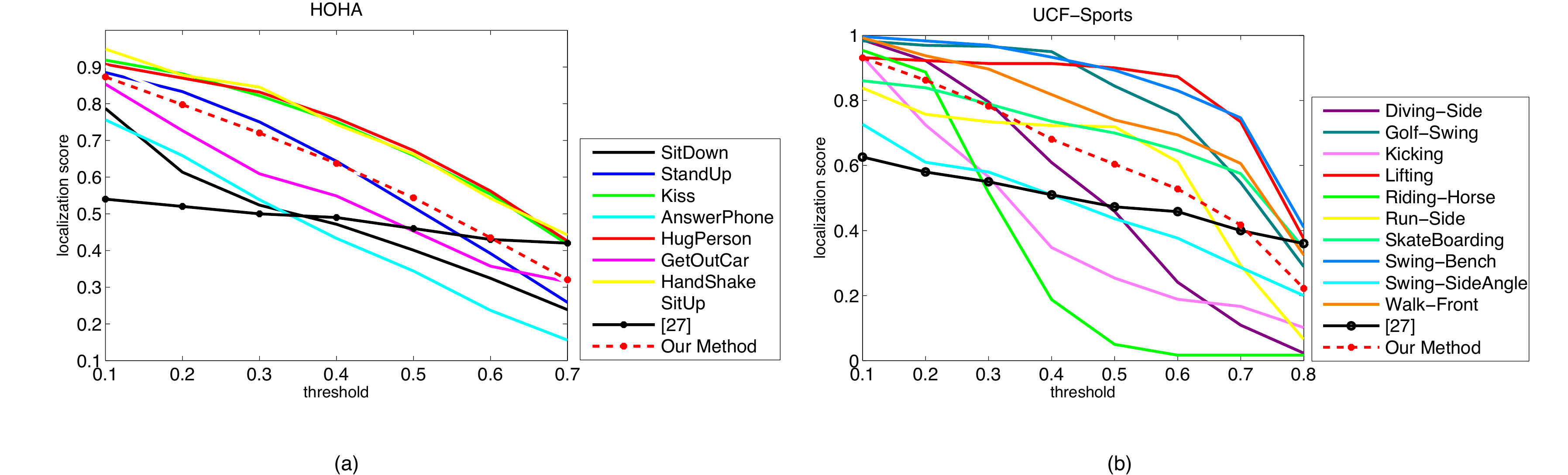}
\caption{\small{Localization scores on HOHA and UCF Sports data set for the our compositional trajectories and the average score of \cite{RaptisSoattoCVPR2012} as function 
of the overlap threshold $\theta$.}}
\label{fig:localucf}
\end{center}
\end{figure}

\begin{table}
\caption{\small{Action recognition performance comparison (on accuracy) on the UCF-Sports dataset with compositional trajectories against the state of the art. All results use training/testing split by \cite{TianMoriICCV2011}. Note that our method dose not require bounding box annotation for action recognition.}}
\label{tb:acrecog}
\centering
\begin{tabular}{|c|c|c|c|c|c|}
\hline
\small{Method} &\small Lan et al. \cite{TianMoriICCV2011}& \small Raptis et 
al. \cite{RaptisSoattoCVPR2012} & \small SDPM \cite{TiSuShCVPR2013} & \small{Ma et al. \cite{MaSclaroffICCV2013}} &\small Our Method\\
\hline
\small{Accuracy}&73.1&79.4&75.2&81.7&78.8\\
\hline
\small{Supervision}&\small{label+box}&\small{label+box}&\small{label+box}&\small{label}&\small{label}\\
\hline
\end{tabular}
\end{table}

\noindent\textbf{Action Recognition}\quad
\label{exp:recognition}
Table \ref{tb:acrecog} compares the average accuracy of our method with results 
reported by other researchers on UCF Sports. Our method performs better than the SDPM \cite{TiSuShCVPR2013} and Lan et al. \cite{TianMoriICCV2011} and is comparable with Raptis et al. \cite{RaptisSoattoCVPR2012}. But, \cite{RaptisSoattoCVPR2012} uses
the bounding box of action in each frame for training whereas our results (and \cite{MaSclaroffICCV2013}) are achieved by only training on the video without 
bounding box. Thus our compositional trajectories can better represent the 
action in the videos. Our accuracy is slightly lower than \cite{MaSclaroffICCV2013} which extracts both static and non-static segments from every frame in a video, probably because our compact representation focuses more on human action and lose some context information.   

\noindent\textbf{Action Localization}\quad
\label{exp:localization}
We compare the performance of our compositional trajectories with 
\cite{RaptisSoattoCVPR2012} in the action localization task. According to the 
evaluation process of \cite{RaptisSoattoCVPR2012}, we obtain the localization 
score for our compositional trajectories.  Fig. \ref{fig:localucf} illustrate the average localization score across the test videos of each action as well as the mean localization score across the two datasets: UCF sports and HOHA. From these figures, we notice that most of our trajectories are inside or around the bounding box of the action (like Fig.  
\ref{fig:visualization} showed for one example). For instance, setting the overlap 
threshold $\theta=0.5$ (which means half of the points in the compositional 
trajectories lie inside the action bounding box at the given frame), we get an 
average localization score of 0.61 and 0.55 for UCF-Sports and HOHA,  
which is significantly better than the localization score in \cite{RaptisSoattoCVPR2012} with 0.473 and 0.484, respectively. This means our compositional trajectories are meaningful for localizing human action as a part of action understanding.

\begin{figure}[t]
\begin{center}
\includegraphics[width=1\linewidth]{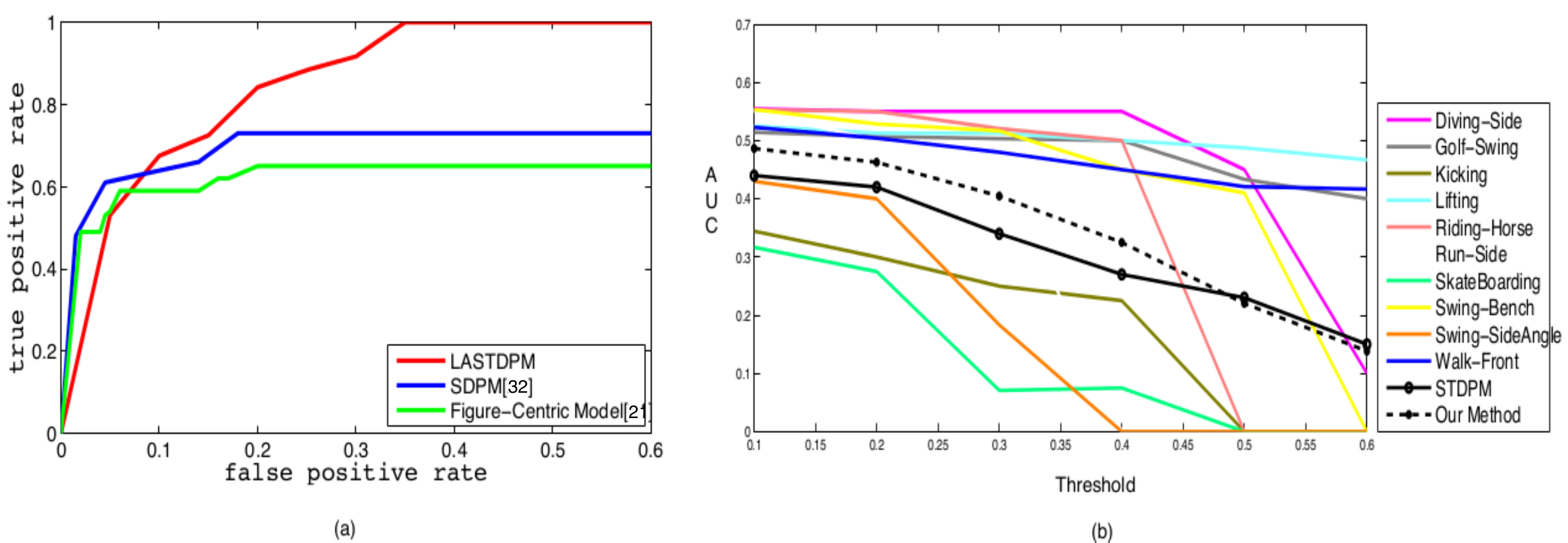}
\caption{\small{Action detection comparisons on UCF Sports. (a) ROC at overlap threshold of 0.2; (b) AUC for threshold from 0.1 to 0.6. The black dot curve shows the average performance of LASTDPM and the black solid curve shows the average performance of SDPM \cite{TiSuShCVPR2013}. Other curves show the detection results for each action by our LASTDPM.}}
\label{fig:detectucf}
\end{center}
\end{figure}

\noindent\textbf{Action Detection}\quad
\label{exp:detection}
We test our new LASTDPM model based on our compositional trajectories on the 
UCF Sports datasets for action detection, we use the standard ``intersection-over-union'' 
measurement, Fig. \ref{fig:detectucf} (a) shows the ROC curve for overlap score of 
0.2; Fig. \ref{fig:detectucf} (b) summarize results (using AUC) for overlap 
scores ranging from 0.1 to 0.6. Clearly, our LASTDPM significantly outperforms 
Lan et al. \cite{TianMoriICCV2011} and SDPM \cite{TiSuShCVPR2013}, which is a two-layer
spatiotemporal deformable parts model based on HOG3D filters.

\subsection{Qualitative Results}
\begin{figure}[t]
\begin{center}
\includegraphics[width = 0.62\textheight]{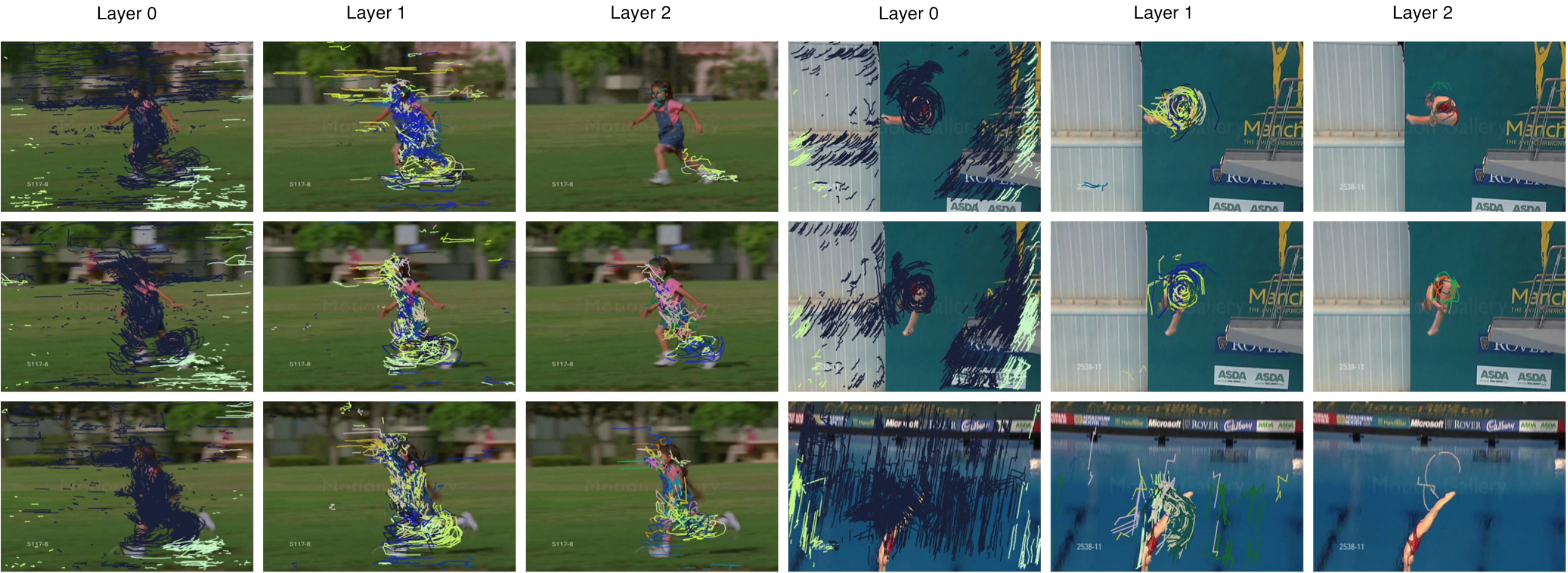}
\caption{\small{Visualization of our compositional trajectories. The columns are sampled frames from two videos of ``Running" and ``Diving-Side" from UCF-Sports data set. Each row shows CT from layer zero, one and two in the hierarchy.}}
\label{fig:visualization}
\end{center}
\end{figure}

We visualize our hierarchies of compositional trajectories, Fig. \ref{fig:visualization} shows elements from layer 0 to layer 2 in sampled frames from  ``Running" and ``Diving-Side" in UCF-Sports data set. Note that elements across frames form the point trajectory and 
we draw the whole trajectory in last frame of the each element. Both videos show our compositional trajectories are able to capture long-ranging human motions, such as the ``curve" of running girl's feet. In addition, our compositional trajectories can effectively remove camera motion because of its lack of statistical significance in the data set, layer 1 and layer 2 of ``Diving" video demonstrate our method successfully keeps human motion and restrain camera motion at the same time.

\section{Conclusion}

In this paper, we view human action as composable elements and propose a compositional structured model for action understanding. First, we propose a new representation called compositional trajectories, which can be used directly in action classification and localization. They also form the feature basis for our locally articulated spatiotemporal deformable parts model that learns the structure of human action with multiple layers of deformable parts to allow for grain-fine articulation. Especially, our subparts can adapt to subtle variation in the way a human may carry out a given task. We implemented both models and test them on three action understanding problems: recognition, localization and detection.  We compare our methods against state of the art approaches on all three problems and find general superior performance. Given the impact the raw dense trajectories have already had to the community, we expect our compositional trajectories to similarly positively impact research in action understanding going forward.

\bibliographystyle{plain}
\bibliography{comptraj_arxiv}
\end{document}